# Neonatal seizure detection from raw multi-channel EEG using a fully convolutional architecture

Alison O'Shea[a,c,*], Gordon Lightbody[a,c], Geraldine Boylan[b,c], Andriy Temko[a,c]

[a] Department of Electrical and Electronic Engineering, University College Cork, Ireland.
[b] Department of Paediatrics and Child Health, University College Cork, Ireland.
[c] INFANT Research Centre, University College Cork, Ireland.

*Corresponding author: alisonoshea@umail.ucc.ie, Department of Electrical and Electronic Engineering, University College Cork, College Road, Cork, Ireland

# Abstract

A deep learning classifier for detecting seizures in neonates is proposed. This architecture is designed to detect seizure events from raw electroencephalogram (EEG) signals as opposed to the state-of-the-art hand engineered feature-based representation employed in traditional machine learning based solutions. The seizure detection system utilises only convolutional layers in order to process the multichannel time domain signal and is designed to exploit the large amount of weakly labelled data in the training stage. The system performance is assessed on a large database of continuous EEG recordings of 834h in duration; this is further validated on a held-out publicly available dataset and compared with two baseline SVM based systems.

The developed system achieves a 56% relative improvement with respect to a feature-based state-of-the art baseline, reaching an AUC of 98.5%; this also compares favourably both in terms of performance and run-time. The effect of varying architectural parameters is thoroughly studied. The performance improvement is achieved through novel architecture design which allows more efficient usage of available training data and end-to-end optimisation from the front-end feature extraction to the back-end classification. The proposed architecture opens new avenues for the application of deep learning to neonatal EEG, where the performance becomes a function of the amount of training data with less dependency on the availability of precise clinical labels.

# 1. Introduction

The detection of seizures in newborn babies is a clinically important task, which has motivated a large body of work in the area of developing and testing automated seizure detection algorithms to support clinical decision making. The presence of seizures in the neonatal period is associated with brain injury; therefore, recognising seizure events when they occur allows medical practitioners to begin treatment promptly, potentially preventing further injury [1]. Seizures in children and adults are typically associated with physical movement or visible indicators. In contrast, most neonatal seizures present no physical signs, especially after anti-epileptic drug use. Research indicates that approximately two thirds of neonatal seizures occur at the sub-clinical level only; therefore, the reliable detection of seizures is only possible through EEG analysis [2]. Applying electrodes to the baby's head and reading the EEG waveforms to identify seizures requires highly trained staff and specialised equipment [3]. In many Neonatal Intensive Care Units (NICU), even when real-time EEG monitoring is available, the availability of on-site expertise to interpret EEG signals is limited and in practice the diagnosis is still based only on clinical signs; reliably detecting as little as 10% of seizure events [2]. These challenges have prompted research into the development of computer-based automated seizure detection algorithms [3]. Automated seizure detection algorithms can raise an alarm and provide objective decision support to clinicians, aiding the prompt detection and treatment of seizures [4], [5].

Early automated seizure detection routines relied on heuristic rules and thresholds. The clinical characteristics of neonatal seizures motivated the search for features that characterise the repetitiveness, order and predictability of ictal EEG. A temporal window of EEG could then be classified as seizure by comparing such features with user defined thresholds, [6], [7], [8]. Three examples of seizure detection algorithms developed in this way were compared by Faul et. al. [9], experiments showed that although each algorithm was effective in detecting certain seizure patterns, the diverse variety of neonatal seizure morphologies meant that none of the algorithm performances were deemed reliable enough for clinical use.

The adoption of more complex data-driven machine learning algorithms followed these simple heuristic classifiers. Increasingly data-driven algorithms were trained for the task of neonatal seizure detection, utilising a similar feature set as outlined in [9], and employing classifiers such as Support Vector Machines (SVM) and Artificial Neural Networks [10], [11]. The majority of these algorithms also required extremely precise seizure labels which require a lot of time, skill and niche expertise to create. In these algorithms handcrafted features were used in conjunction with a data-driven machine-learning based decision making. The handcrafted features, on which the approaches depend, are reliant on prior knowledge of neonatal EEG. The features are usually extracted from the time, frequency and information theory domains to provide energy, frequency, temporal and structural descriptors of the neonatal EEG, giving an informative characterization of each segment of EEG [10], [12].

One such algorithm combines many simple features into a powerful classification framework [4]. Other works have relied on the identification of increasingly complex features to improve system performance. Complex representations such as chaos theory and time-frequency analysis [13], [14], have been explored in an effort to find a single feature which can give separation between seizure and non-seizure classes. A combination of complex features could be more representative of the natural interdependencies in EEG signals. These more complex features are often well correlated with seizures in neonatal EEG, but they can increase the time required to complete the feature extraction stage.

Progress in deep learning-based research in domains such as image and audio processing has been utilised in the development of new EEG classification algorithms that do not require a feature extraction stage. These, deep learning models can take temporal or spectral EEG as input and using back propagation, they can learn both feature extraction and segment classification routines in one end-to-end optimisation procedure [15]. Deep learning algorithms have been applied to paediatric and adult EEG for a variety of different tasks, such as brain computer interfaces, seizure detection and feature extraction [16], [17], [18], [19], [20], [21]. Limited work has been carried out utilising deep learning for the neonatal seizure detection task.

A recent algorithm was developed for the neonatal seizure detection task using a Convolutional Neural Network (CNN) as a feature extraction stage followed by a decision tree as a classifier [22]. The intermediate output from the CNN is used as input to a decision tree classifier, this is an ensemble technique called stacking which is employed to boost the performance. In this type of algorithm, the CNN is a complex feature extractor and the decision tree is a classification algorithm. This hybrid model is different to the end-to-end deep learning framework developed in this work.

The noisy and artifact prone nature of EEG means that decomposed representations are more commonly used as input to deep learning algorithms. These decompositions usually take the form of Fourier or wavelet transforms; deep learning algorithms are assumed to more easily extract relevant features from these representations. Wulsin et. al. [23] have performed experiments comparing the performance of a deep belief net classifier for detecting anomalies in EEG waveforms; one network was trained with domain relevant features as input, the other with the raw EEG data. The architecture trained using raw input gave better performance [23]. In other experimental setups in the same work, the use of raw EEG as the classifier input gave no reduction in performance. Previous publications in the area of neonatal seizure detection using EEG have proven the robustness [15], and computational efficiency [24] of training deep learning algorithms with raw time-series inputs.

Training deep learning algorithms using raw neonatal EEG signals is a crucial new step forward in the development of neonatal seizure detectors that rely on EEG waveforms to locate seizure events rather than human-extracted features, which have the potential to introduce biases. Learning from raw data is especially suited for EEG processing tasks, where the inherent complexity of the EEG signal means that not all information can be reliably recorded in an a priori extracted feature set.

Deep learning presents an important new approach to signal processing challenges, but there are some important drawbacks associated with deep learning algorithms. For effective training a large training dataset is required, which is usually orders of magnitude larger than a dataset that would be required to train a shallow machine learning algorithm. The training stage in a deep learning algorithm requires many iterations through the training dataset, which means that efficient training requires large computational resources. Deep learning algorithms are typically considered to be less interpretable than other machine learning algorithms.

The move towards deep learning has solidified the split between two branches of research in the area of neonatal seizure detection, the areas of hand-crafted feature engineering and advanced classifier design. This new entirely data-driven world of deep algorithms is developing separately to the world of feature engineering, which is seeing the development of more complex, task specific features [13]. The classifiers discussed and developed in this work are pushing towards the goal of an entirely data-driven classifier where the algorithm has a single end-to-end optimisation routine.

This study has developed deep learning algorithms for the neonatal seizure detection task, with the focus of fully exploiting the benefits of using CNNs. In this work, the EEG signals are represented in the time domain, therefore temporal EEG signals are used for training and testing the deep learning algorithms. Two feature based machine learning algorithms will be used as baselines in this work; one baseline relies on complex and highly-engineered features, the other uses a large set of many simple features [13], [25]. These baselines require seizure annotations which are precisely located in both time and space. This paper presents a CNN solution which can be trained using weak seizure labels that require a lot less clinical labour to obtain. Section 2 of this paper outlines the datasets used in the work and the two algorithms that will be used as performance baselines for these datasets. Section 2 also shows the methodology used to develop, select and test the deep learning architectures. Section 3 shows the results of experimental work including the comparison of this fully data-driven work with two baselines. Section 4 discusses the presented results, with conclusions drawn in Section 5.

# 2. Methods

## 2.1 Datasets

### 2.1.1 Dataset I – Cork

A dataset of EEG signals recorded in the NICU of Cork University Maternity Hospital, Ireland was used in this work. The Cork dataset includes recordings from 18 full-term newborn babies, between 39 and 42 weeks gestational age (GA), all of whom experienced seizures because of brain injury associated with hypoxic ischemic encephalopathy (HIE). The Cork dataset totals over 834 hours in duration and comprises of 1389 seizure events. More details about the seizure events can be seen in Table 1. All files contain eight channels of EEG, recorded using the 10-20 placement system modified for neonates. A bipolar montage of the following eight channels was used: F4-C4, C4-O2, F3-C3, C3-O1, T4-C4, C4-Cz, Cz-C3, C3-T3. Raw multi-channel EEG data were collected using the Carefusion NicOne video EEG monitor at a sampling rate of 256Hz.

Two independent neonatal electro-encephalographers annotated all seizure events in this dataset. The agreement of these experts was taken as the ground truth and any disagreements were discussed until consensus was reached by both experts. The experts had the EEG waveforms and a synchronised video recording of the cot available to them while annotating the seizure events. No data pre-selection or artifact rejection through engineered rules was performed, to avoid human-induced biases. The artifacts which are normally present in long-term EEG will form part of the data-driven training process. The data are reflective of a real-world NICU environment where EEG signals would contain a variety of different artifacts. The SVM based neonatal seizure detection algorithm developed in [25] was trained and tested using the same dataset; the published performance of that algorithm is directly comparable with this work. This study had full ethical approval from the Clinical Research Ethics Committee of the Cork Teaching Hospitals.

### 2.1.2 Dataset II – Helsinki

A publicly available dataset of annotated neonatal EEG waveforms collected at Helsinki University Hospital is also used as a test dataset in this work [26]. This publicly available test dataset will be referred to as the Helsinki dataset. This dataset contains 18-channel EEG segments from 79 full-term babies; signals are sampled at 256Hz. The dataset is not composed of long continuous unedited recordings but rather contains 1-2h excerpts per baby. As a result, in contrast to the Cork dataset, this dataset is larger in the number of babies (79 vs 18 in Cork) but much smaller in both the overall duration (112 hours of recordings vs 834 hours in Cork), and in the number of seizure events (342 vs 1389 in Cork). The seizure annotations were defined based on the consensus of three EEG experts. Out of 79 babies, 39 experienced seizures.

*Table 1 The Cork dataset of full-term neonatal EEGs with seizure annotations.*

| Baby Number | Record length (h) | Seizure events | Mean seizure duration | Min seizure duration | Max seizure duration |
| --- | --- | --- | --- | --- | --- |
| 1 | 29.66 | 17 | 1'29'' | 17'' | 3'54'' |
| 2 | 24.74 | 3 | 6'10'' | 55'' | 11'9'' |
| 3 | 31.43 | 209 | 1'50'' | 11'' | 10'43'' |
| 4 | 47.54 | 84 | 1'37'' | 32'' | 9'58'' |
| 5 | 56.78 | 63 | 6'32'' | 20'' | 34'10'' |
| 6 | 19.24 | 46 | 1'7'' | 15'' | 4'17'' |
| 7 | 60.75 | 99 | 1'31'' | 14'' | 10'20'' |
| 8 | 49.55 | 17 | 5'55'' | 29'' | 19'14'' |
| 9 | 67.76 | 201 | 4'59'' | 13'' | 37'6'' |
| 10 | 59.79 | 41 | 4'51'' | 13'' | 34'46'' |
| 11 | 22.21 | 43 | 2'27'' | 17'' | 7'36'' |
| 12 | 54.36 | 150 | 1'35'' | 15'' | 10'8'' |
| 13 | 51.68 | 60 | 3'26'' | 19'' | 16'56'' |
| 14 | 22.79 | 21 | 8'12'' | 22'' | 39'3'' |
| 15 | 66.02 | 121 | 1'30'' | 10'' | 7'8'' |
| 16 | 76.39 | 189 | 5'4'' | 26'' | 34'37'' |
| 17 | 30.7 | 21 | 5'30'' | 27'' | 23'16'' |
| 18 | 63.02 | 4 | 9'33'' | 7'19'' | 13'22'' |
| Total | 834.41 | 1389 | | | |

### 2.1.3 Neonatal seizure annotations: weak and strong labels

Both the Cork and the Helsinki datasets have seizure annotations. Seizures can be annotated in different ways. First, the babies can be annotated as having experienced seizures; this differentiates them from babies who do not have seizures. While these labels are clinically valuable, they provide no spatial or temporal information about the seizures and are not suitable for developing automatic seizure detection systems. Second, seizures can be localised in time with overall annotations, whereby the start and end time for each seizure event is recorded for each file giving precise temporal resolution to each seizure event - this is a weak label. Figure 1 (a) shows weak annotations for a seizure, where the start and end times of the seizure are indicated, but the channels involved in the seizure event are not specified.

Propagating these overall annotations to each EEG channel will introduce a lot of noise, since seizures can be localised to a single channel and even migrate from one channel to another during the seizure event. These overall annotations are sufficient for evaluation of the algorithmic performance but are not enough for the extraction of seizure signal signatures to train a classifier. In this work, we refer to these annotations as weak annotations.

Thirdly, in addition to overall annotations, seizures can be localised in space, whereby each seizure (or a part of it) has not only its start and end time but also its spatial location specified as shown in Figure 1 (b). We will refer to these as strong labels. There is a large cost associated with the acquisition of strong labels as it requires a lot of extra time and effort from clinical staff. In the datasets of neonatal seizures used in this work a small minority of the seizure events have associated strong labels, due to the large cost associated with acquiring strong labels.

The target when designing a neonatal seizure detection algorithm is to perform well on EEG data of unseen newborn babies. This is different from seizure detection and prediction in adults where patient specific systems can be of interest [27]. The generalisation requirement prevents engineers from concatenating the EEG features from different channels and using the overall annotations to train the classifier. If EEG features are concatenated across channels this will lead to the classifier learning seizure-location dependencies, which are too baby specific and will not allow the algorithm to generalise to new data. Theoretically, if a huge dataset is used for training it could cover all possible combinations of spatial seizure locations. However, in reality datasets of this size are not available, and this is not a good use of the limited data that are available. Thus, in order to train a classifier, a single EEG channel is processed at a time in the state-of-the-art algorithms [12], [13]. In order to train single channel classifiers, the location of the seizure must be available i.e. the seizure labels must be strong.

Seizures that happen in a single channel and seizures that happen in multiple channels are treated with equal clinical significance. The information on the spatial location of the seizure is in general not available. Acquiring channel-specific seizure annotations creates a significant extra burden for the neonatal neurophysiologist who is labelling the data. While all seizure events in the Cork and Helsinki datasets have weak labels, in the Cork dataset, a minority of seizure events have channel-specific annotations; out of 4503 minutes of seizure activity, only 617 minutes have channel-specific annotations, this was an exceptional additional workload for the clinical annotators involved. For the Helsinki dataset, the strong labels are also publicly available. It is worth mentioning that only weak labels are required for evaluation of the neonatal seizure detection algorithms, because the algorithm is only required to output the temporal location of seizure events.

### 2.1.4 Neonatal seizures

Background EEG of both healthy and sick newborns appears random in nature [28]. In contrast to this pseudo-random background behaviour, seizures in neonatal EEG are defined by a repetitive, evolving pattern, the voltage must be larger than 2µV and the pattern must be longer than ten seconds in duration for an event to be classified as a seizure [29]. An example of a seizure in a newborn can be seen in the grey shaded regions (strong labels) of Figure 1(b).

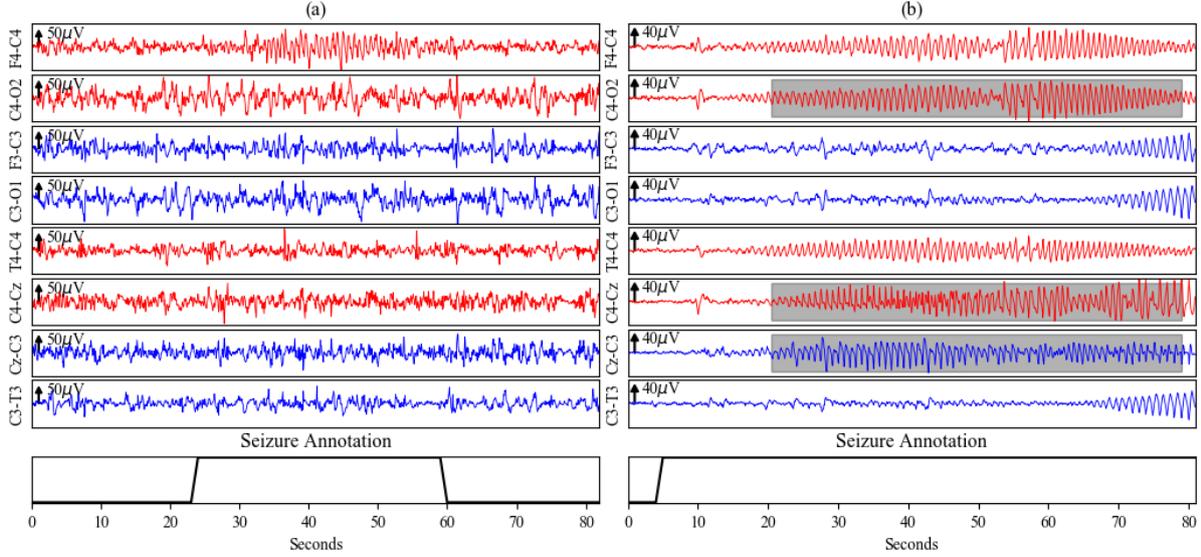

*Figure 1 Examples of EEG showing a neonatal seizure annotated with weak (a) and strong labels (b). Channel labels are shown on the horizontal axis. It can be seen that a weakly labled seizure (a) is observable only in the F4-C4 channel, which cannot be derived from the weak label (shown below the EEG trace). The strong labels in (b) uniquely localise the seizure activity (shown as grey shading). However, this strong label does not imply that these are the only channels where the seizure is present, but these are the only channels where a strong label is available.*

In order to detect a seizure, the neurophysiologist looks for a rhythmic and evolving pattern that deviates from the background activity. Seizures can evolve spatially by moving across different brain regions, they also evolve in frequency over time. Seizure burden is a measure of the percentage of time spent in a seizure state. A higher seizure burden in the neonatal stage is associated with more adverse outcomes [29].

These rhythmic patterns have prompted the search for features which can measure repetitiveness and find patterns which contrast with the random background EEG. However, the EEG electrodes will also record artifacts that possess a similar rhythmic structure, such as the ECG rhythm, movement, sweat or the baby's respiration.

## 2.2 Machine learning approaches

This section describes two feature-based machine learning baselines. The first baseline uses a large feature set of relatively simple EEG features, the algorithm is data-driven as these relatively simple features are used in conjunction with an SVM algorithm to learn a seizure detection routine. The second baseline relies on more complex features which were engineered to recognise seizures in EEG, this algorithm utilises the power of hand-crafted feature engineering and also leverages an SVM algorithm to detect seizures in EEG. Both feature-based and data-driven machine learning solutions rely on extraction of the seizure signature from a single channel of the EEG signal [12], [13], [15]. This imposes the need for strong labels during the training stage i.e. seizure events that are annotated both in time and in space.

### 2.2.1 SVM Cork

This algorithm is based on an SVM classifier and was evaluated here using the same procedure and the same Cork dataset as used in [25]. In the pre-processing stage, the raw EEG sampled at 256Hz is band-pass filtered between 0.5Hz and 12.8Hz, down-sampled to 32Hz and split into 8-second windows with a 4-second shift. The SVM algorithm relies on 55 features extracted from a moving 8-second window of EEG. The features were carefully engineered for the seizure detection task. A full list of features is reported in [12] with the github repository link for the implementation available in [27]. A Gaussian kernel SVM was trained on the extracted features. This training framework requires strong labels in order to create a representation of the seizure data which avoids patient specific seizure location dependencies. In the testing stage, which is shown in Figure 2 (left path), the trained SVM model is applied to each EEG channel separately with the final decision made after post-processing and fusing the probabilities across all EEG channels. The post-processing stage includes taking a maximum probability across channels, a moving average filter of 1-minute length, adaptation to the background level of probability and a collar to compensate for the effect of smoothing. More details on this data-driven baseline system can be found in [25].

One key takeaway from this work is that when many relatively simple EEG signal features are combined with a powerful and robust classifier, such as the SVM, a reliable detector can be trained. This algorithm has been successfully validated in a clinical trial in eight NICUs across Europe [30]. This algorithm's performance is used as a baseline on the Cork dataset in this work.

### 2.2.2 SVM Helsinki

The algorithm, referred to as SVM Helsinki in this work, was trained on the publicly available Helsinki dataset [13]. This SVM based algorithm uses 21 features extracted from a moving 32-second window of EEG, with a 4-second shift. These features are derived from the time domain and frequency domain, and additional bespoke features based on time-frequency correlations are defined and included. The paper focuses on the design of these new features, which were developed by the authors to specifically fit the neonatal seizure detection task.

The SVM Helsinki algorithm is an example of the power of feature engineering, the authors show that the developed features are well correlated with neonatal seizures. These features are more computationally intensive than the set used in the SVM Cork. The SVM Helsinki algorithm also proposes using different pre-processing routines for different features. The performance on the publicly available feature-driven Helsinki dataset has been published [13], and the code necessary to implement the algorithm is available online [31]. This algorithm's performance is used as a baseline on the Helsinki dataset in this work.

### 2.2.3 Comparison of SVM algorithms

Both SVM algorithms extract a set of meaningful features from each of the EEG channels, the classifiers are applied to each EEG channel separately, and the per-channel probabilistic decisions are combined in the post-processing stage. This routine means that both algorithms require channel-specific seizure annotations for training.

The Cork algorithm uses a large set of simple EEG features. The Helsinki algorithm uses a smaller set of features, but two of these features are based on spike correlations and time-frequency correlations, which are known to be good predictors of neonatal seizures, making it an example of a feature driven algorithm. These two highly engineered correlation features were informed by both engineering and medical prior knowledge. The features used in the Helsinki SVM algorithm are tuned to discriminate seizures from background EEG, but they require considerable computational power to classify a single window of EEG.

The Cork algorithm was developed using the Cork dataset, consisting of long EEG recordings and a seizure burden of around 10%. The post-processing includes a 60-second moving average filter, a collar operation (the collar length was set at half the length of the moving average filter) and a respiration artifact adaptation algorithm. The respiration artefact adaption algorithm is designed to adaptively compute the seizure probability with respect to the level of background artifact corruption. This makes the system more robust to long-term artifacts, which are typically found in the long unedited EEG recordings that are collected in the NICU.

The Helsinki algorithm was developed using the Helsinki dataset, consisting of short EEG recordings and a seizure burden of around 20%. The post-processing includes smoothing and decision making on a 48-second window which was selected to maximise the performance on the dataset.

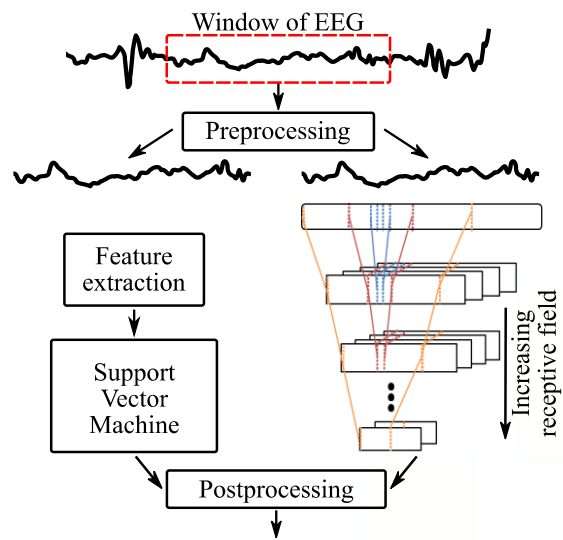

*Figure 2 A comparison of the SVM algorithms (left branch) and the deep learning architecture (right branch) developed in this work [39]. The algorithms process windows of EEG. In this work, the post-processing and pre-processing routines utilised in the SVM Cork are applied. The CNN directly substitutes for the feature extraction and SVM blocks in the baseline SVM algorithms.*

## 2.3 Deep learning for neonatal seizure detection

The seizure detection system in this work is the first of its kind, as it is the first end-to-end deep learning system designed for the task of neonatal seizure detection. In this work, the network inputs are temporal EEG waveforms; this may not be the most obvious choice and many different input representations were tested in the process of developing this classification routine. Algorithms in Section 2.2.1 and Section 2.2.2 use features to represent the EEG within a specified moving window. Many features rely on Fourier transform representations; this is an issue when working with EEG, as signal stationarity cannot be assumed, even over small intervals. The purpose of experimenting with deep learning architectures was to circumvent the need for feature extraction, to input the entire signal into a classification algorithm instead of a human-defined set of features.

In initial experiments, the EEG was converted to image representations and a version of the architecture with 2D convolutional filters was tested. Experiments were conducted using the spectrogram of the EEG, but results were inferior to previous baselines on the same task. In order to focus the learning on the lower frequency ranges within the EEG signal the problem was reformulated as one of learning the filter-bank weights of the spectral envelope. The aim was to produce a frequency scaling that would be similar in concept to Mel-scaling, which is used in speech processing [32]. Some improvement was obtained with respect to the previous experiment, but the performance was still inferior to that of the baseline. Subsequent experiments used 1D convolutional filters applied to raw EEG inputs, the filters were selected to be wide; 64 or 128 samples long, which corresponds to 2 or 4 seconds (of pre-processed signal). In these experiments, the networks gave performance improvements over the architectures that used image representations as input; the networks with raw waveform inputs showed the ability to learn the underlying characteristics of EEG. Finally, sample size filters, which are an important feature of the architecture used in this work, were tested. In contrast to larger filter lengths, sample-sized filters allow for the learning of the EEG filters in a hierarchical manner [33]. This work uses filters that are only three samples wide in all deep learning architectures; this corresponds to less than one tenth of a second in the first convolutional layer.

### 2.3.1 Building blocks of the fully convolutional architecture

In order to create the deep learning architecture for neonatal seizure detection, the feature extraction and classification blocks need to be defined using deep learning layers. Figure 3 shows the structure of these blocks; it should be noted that the only layers required to implement these blocks are convolutional, normalisation, activation and pooling layers. Figure 3 (a) is a block which forms the feature extraction section of the deep CNN architecture. These blocks can be stacked to increase network depth. Figure 3 (b) shows the architecture of the classification block which allows the learned intermediate representations of EEG to be classified as either seizure or non-seizure (background).

In a typical CNN architecture, the classification stage is implemented using densely connected layers. In the fully convolutional network (FCN) architecture convolutional layers are used both to extract features from inputs and to perform classification, as illustrated in Figure 3. The number of filters in the classification block is chosen to be equal to the number of classes. A global average pooling is calculated across each of the feature maps in the final layer, this is followed by a softmax activation function that converts the values into a set of probabilistic outputs. In this manner, since there are no fully connected layers, the network is optimised to extract the necessary discriminative information through the hierarchy of convolutional filters before reaching global average pooling, thus learning the 'seizure-ness' and 'background-ness' of the input raw EEG.

### 2.3.2 Convolutions as data-driven feature extraction

All of the convolutional and pooling filters are 1D. The output of each convolutional layer is of shape ($K$, $I$), where $K$ is the number of feature maps and $I$ is the width of each feature map. The $I$ parameter is directly related to the time axis at the input layer; temporal ordering in the $I$ axis is maintained until the final global pooling layer. The convolutional mapping between layer input $x$ and layer output $h$ can be seen here:

$$h_i^k = \max\left(0, ((W^k * x)_i + b_k)\right), \tag{1}$$

where $W$ is the weight matrix. $(W^k * x)_i$ refers to convolutional calculation, at position $i$, for the $k^{th}$ feature map, taking inputs from all feature maps in the previous layer. $b_k$ is a bias added to all calculations in the $k^{th}$ feature map. The ReLu non-linear operation is applied to the output of the convolutional operation.

### 2.3.3 Parameters

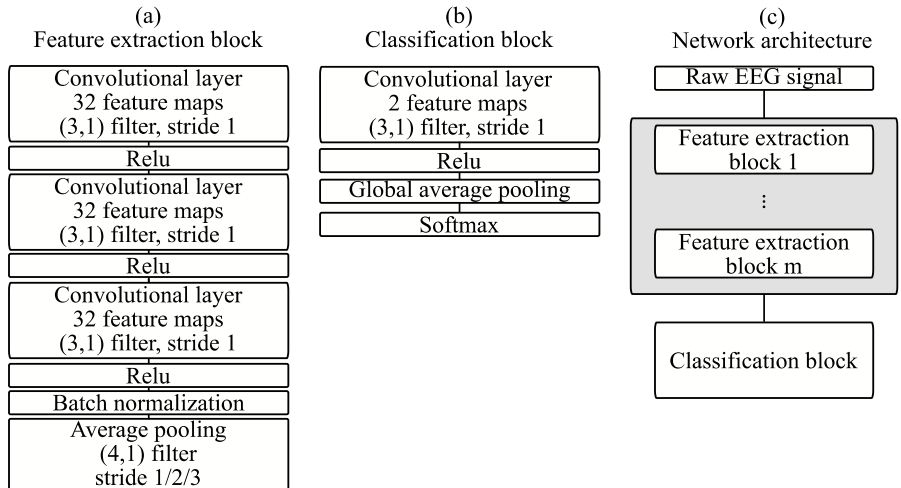

Figure 3 The structure of each feature extraction block (a) and classification block (b) used in experiments. The full network architecture is shown in (c). Varying the number of feature extraction blocks (j) and the pool stride value controls the network complexity and the receptive field in the final convolutional layer.

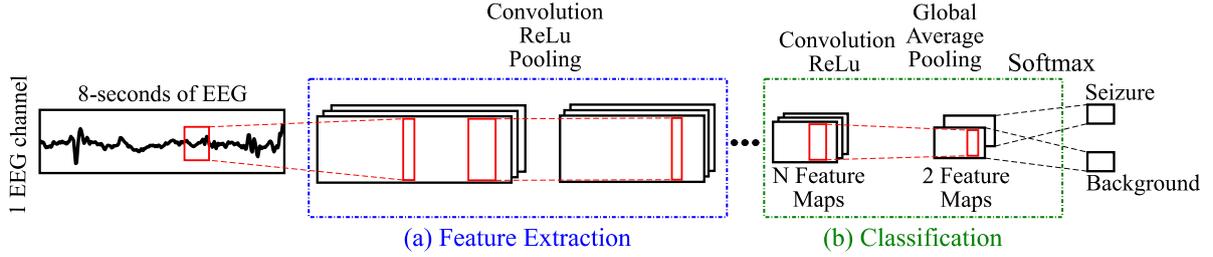

*Figure 4 The **1D FCN** designed for the seizure detection task. This network relies only on convolutional, pooling and batch normalisation layers. The convolutional layers make up the feature extraction block (a) of the network; they can have any amount of feature maps; in this work the number of feature maps is always 32. The final convolutional layer, which learns to classify the inputs into seizure or background categories, has the same number of feature maps as network output categories (b). After a global pooling layer, these final feature maps connect directly to the individual network outputs.*

In this work, network depth can then be arbitrarily chosen by varying the number of feature extraction blocks Figure 3 (c). More blocks in the network allows for the learning of deeper networks with increased feature complexity. The pool stride in each feature extraction block refers to the sample stride between successive pooling operations within the feature extraction blocks. Increasing the stride has a down-sampling effect, reducing the number of samples in subsequent layers.

A convolutional filter in each convolutional layer is influenced by a certain span of EEG samples in the input layer. The receptive field is the range of samples in the input signal corresponding to the current convolutional weight. In the first convolutional operation, the receptive field is equal to the size of the weight matrix, but in deeper layers the receptive field widens. The range of samples in the input layer that is seen by a filter can be increased by adding convolutional layers or by down-sampling using pooling layers. The maximum possible receptive field in the last layer in these experiments is 256 samples of EEG (8-seconds at a sampling rate of 32Hz). The 8-second window size with 1-second shift was chosen based on prior work, and it is maintained at this value to ensure the comparability with previous work [15].

The receptive field (RF) in each layer can be computed recursively as:

$$RF_M = (RF_{M-1} - 1) \times s + f, \qquad (2)$$

where layer $M$-1 precedes layer $M$, s is the stride and $f$ is the filter width of either a convolutional or pooling layer. In this work, the pool stride takes values of one, two and three samples. From Equation (2), the filter stride has a multiplicative influence on the receptive field width, while the filter width has an additive influence.

In CNN architectures employing fully connected layers, the value of the receptive field in the final convolutional layers is not critical; the presence of dense connections is the equivalent of having a receptive field that spans the entire input. However, in the FCN architecture proposed in this paper, the final convolutional layer has the widest receptive field, these convolutions are the most complex and they span the greatest number of input samples. Thus, the final convolutional layer acts as the classification layer and its receptive field length determines how much of the input EEG is seen by each element before global average pooling.

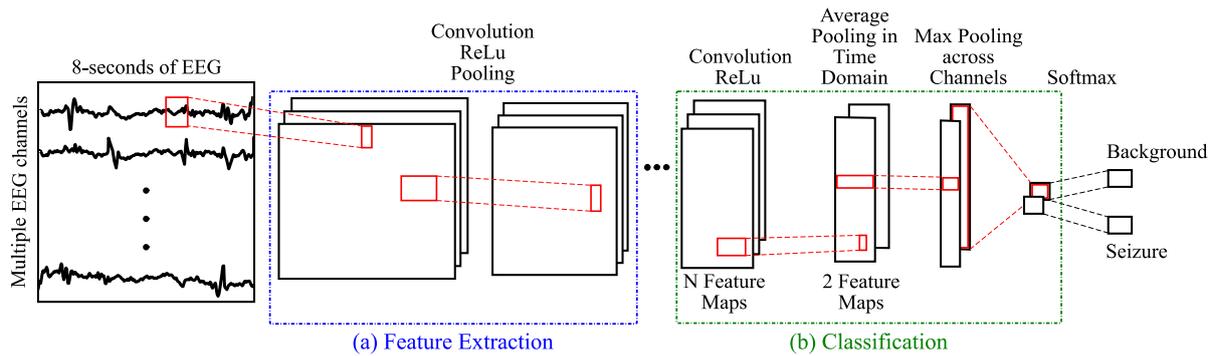

*Figure 5 The **2D FCN** designed for the seizure detection task. This network is designed to take multiple channels of EEG as input and it outputs the seizure decision across all channels. An average pooling layer is designed to create a value which corresponds to the level of seizure-ness in each channel and a maximum pooling layer recognises if there is a seizure in any one channel.*

## 2.4 1D FCN architecture: deep learning baseline

The first network in this work utilises a 1D FCN architecture which is presented in Figure 4; this also illustrates how the final convolutional layers in the FCN are designed to act as classification layers. The EEG input is represented as a 1D temporal waveform. The input is an 8-second window of EEG, sampled at 32Hz, resulting in a 256x1 input vector. Figure 2 (right path) positions the designed 1D FCN architecture with respect to the SVM-based system [25] – similarly, the 1D FCN operates on a single EEG channel and shares the same pre and post-processing steps but it combines both the feature extraction and classification blocks into one end-to-end optimisation routine. Similar to [12], [13], [15], [25], the 1D FCN architecture requires strong labels for training. The usage of strong labels implies that only a small sub-set of the available data can be used for training.

## 2.5 2D FCN architecture: levering more data by using weak labels

An alternative deep learning architecture is also explored in this work, which is shown in Figure 5. This architecture is a 2D FCN that takes multiple EEG channels as input and detects if a seizure happens in any one of the channels. The input can take an arbitrary number of EEG channels. In this work, each input channel is an 8-second window of EEG sampled at 32Hz, resulting in a 256-by-*N* array, where *N* is the number of input channels. The main architectural difference with respect to the 1D FCN architecture is the inclusion of a final maximum pooling layer, which takes the maximum across feature maps from all EEG channels. This max pooling layer allows one of the post-processing steps required in the 1D FCN to be incorporated into the end-to-end learning routine.

The 2D FCN architecture is designed using the same layers as the 1D FCN network with all filters being one dimensional, which implies that no inter-channel correlations are considered to avoid learning baby specific seizure location dependencies. Instead, this architecture is designed to learn the presence of seizure among all available channels, regardless of the seizure location. The 2D FCN network now must accomplish a much more complex task – to learn to distinguish between seizure activity and non-seizure activity from weakly labelled data. Weak labels may have up to 1-to-8 signal-to-noise ratio – e.g. when a seizure is only present in a single channel out of 8 EEG channels as shown in Figure 1 (a). All 8 channels are presented to the network and the same shared convolutional filter will be tuned with EEG samples from all signals. The main advantage of this architecture is that it does not require the seizure labels to be channel-specific; this means that the entire Cork dataset could be used for training instead of the subset that has channel-specific labels. The hypothesis is whether the availability of over 40 times more training data (in terms of the number of EEG samples) can compensate for using noisier weak labels.

## 2.6 Experimental procedure

### 2.6.1 Leave one out experiment

Classifiers need to be validated in a patient independent framework in order to properly assess their ability to perform on unseen data. The leave-one-out cross-validation (LOO) method is a suitable test procedure for a patient independent system. In this method all babies except one are used for training, the trained classifier is then tested by checking its performance on the one unseen baby. This routine repeats until every baby has been treated as the held-out test baby. All experiments follow the same training routine.

The need for strong labels in the 1D FCN imposes limitations on how much of the dataset can be used for training. Since only a small portion of the data are used for training, the remaining EEG from the 17 training babies can be used in the validation set for early stopping, to prevent over-fitting. When the validation score stops improving for more than 25 iterations the training stops. The network architecture that gave the best validation score is then used for testing. The optimization routine uses stochastic gradient descent with a learning rate of 0.001 and Nesterov momentum set to 0.9. The training batch size is 4096. Each training example is a 256-by-1 vector, which allows the network to be trained with a relatively large batch size when compared to image processing examples where the inputs are 2D images with a further colour (red-green-blue) dimension [34]. The loss function is categorical cross-entropy.

In experiments with the 2D FCN architecture, the whole duration of 17 training babies can be used for training, since all of the data have weak labels. In this process (while one baby is used for testing, as per the LOO routine), 3 babies out of 17 are used for validation and early stopping, the remaining 14 babies are used for network training. To balance the target classes, the seizure and non-seizure training examples are weighted in the loss function during training. Since the division of training data into train and validate sets is no longer deterministic, the split is repeated three times by randomly choosing three validation babies from 17. In each split, when the validation score stops improving for more than eight iterations the training stops. The network architecture that gave the best validation score is used. The probabilistic outputs from 3 trained networks are averaged to calculate the test probabilities. The optimization routine uses stochastic gradient descent with a learning rate of 0.001 and Nesterov momentum set to 0.9. The training batch size is 300. Each training example is a 256-by-8 array; the training batch size was therefore reduced to account for the increased input size. The loss function is binary cross-entropy. The Keras TimeSeriesGenerator function was used to transform the multi-channel EEG data into training samples for the 2D FCN.

### 2.6.2 Public dataset and algorithm

A once off test on a publicly available dataset is used to understand the generalisation ability of the trained network. Two algorithms trained on the entire Cork dataset, one using the 1D architecture and the second using the 2D architecture, are tested on the unseen Helsinki dataset. The performances of the deep learning networks on the Helsinki dataset are compared with that of the algorithm referred to as SVM Helsinki in this work; this algorithm was trained on the publicly available Helsinki dataset.

The code required to implement the Helsinki SVM algorithm is available online, therefore the performance of this algorithm in a once-off test of the Cork dataset is reported.

### 2.6.3 Metrics

In this work, the area under the receiver-operating curve (AUC) is the primary measure of classifier performance. The curve is a plot of sensitivity versus specificity; the closer the AUC is to unity the better the classifier's performance. In this task, sensitivity corresponds to the percentage of correctly labelled input windows containing seizures. In this study the relative performance of a deep learning based classifier is reported on two datasets which have baseline performances available in literature and were assessed using the AUC metric [13], [25]. The use of AUC as a metric in this work allows for direct comparison with these performance baselines.

For a seizure detection system to be clinically usable, the specificity needs to be high to result in an acceptable number of false detections per hour. The AUC90 metric is the area under the curve for specificities over 90%. This is a more clinically relevant performance measure, as it looks at the amount of seizure epochs detected for low levels of false seizure detections; low specificity would result in unwanted false alarms in the NCIU.

The AUC metric can be calculated per baby and then averaged or computed across the whole dataset where all recordings are first concatenated. By using the AUC concatenated metric across all babies the babies who do not have annotated seizures can be included in the AUC score. This can be used to report the performance of the classifiers on the Helsinki dataset, as in this particular dataset there are 39 babies who experienced seizures and 40 babies who did not.

The performance baselines used in this work both report high AUC values, which leaves little room for improvement. The relative improvement shows how much the algorithm developed in this work improves with respect to the potential for improvement. At AUC values close to 100% even small improvements in AUC values could lead to superior performance in terms of clinical event-based metrics such as the number of false seizure detections per hour [35].

## 3. Results

Figure 6 shows a comparison between the Cork SVM algorithm and the 1D FCN deep learning algorithm for the LOO experiments for each of the 18 babies in the Cork dataset. The SVM algorithm results presented were reported in [25]. In order to ensure a direct performance comparison, the same pre-processing filters, 8-second windowing and post-processing steps are applied to both algorithms. Both algorithms were trained using the subset of channel-specific annotated seizures in the Cork dataset. In almost every baby the AUC score was improved using the deep learning approach, the average across all AUC scores is improved by over 1%, absolute when using the 1D FCN algorithm.

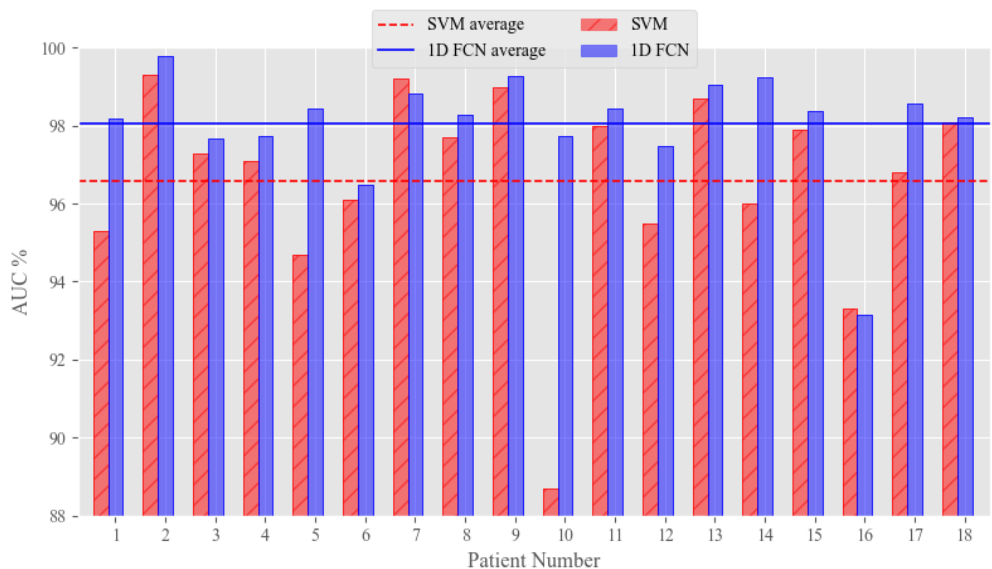

*Figure 6 A comparison of LOO results for each of the 18 babies in the Cork dataset. Results of the previous state-of-the-art for this dataset, an SVM based architecture which uses 55 features extracted from EEG are reported alongside the results of a deep learning model trained using the same set of per-channel annotations. Horizontal lines show the average performance across the 18 babies for the CNN and SVM algorithms.*

Figure 7 shows how test (a) and validation (b) performances in the LOO experiments vary with different parameter combinations. The two variables in these experiments are the number of feature extraction blocks used, defined in Figure 3 (a), and the stride of the pooling operations in the feature extraction blocks. The horizontal axes represent the experiments with the number of feature extraction blocks gradually increased from 1 to 5. The shallowest architecture implemented in this work uses one convolutional block followed by one classification block (4 convolutional layers in total). The deepest architecture used in experiments has five feature extraction blocks and one classification block (16 convolutional layers in total).

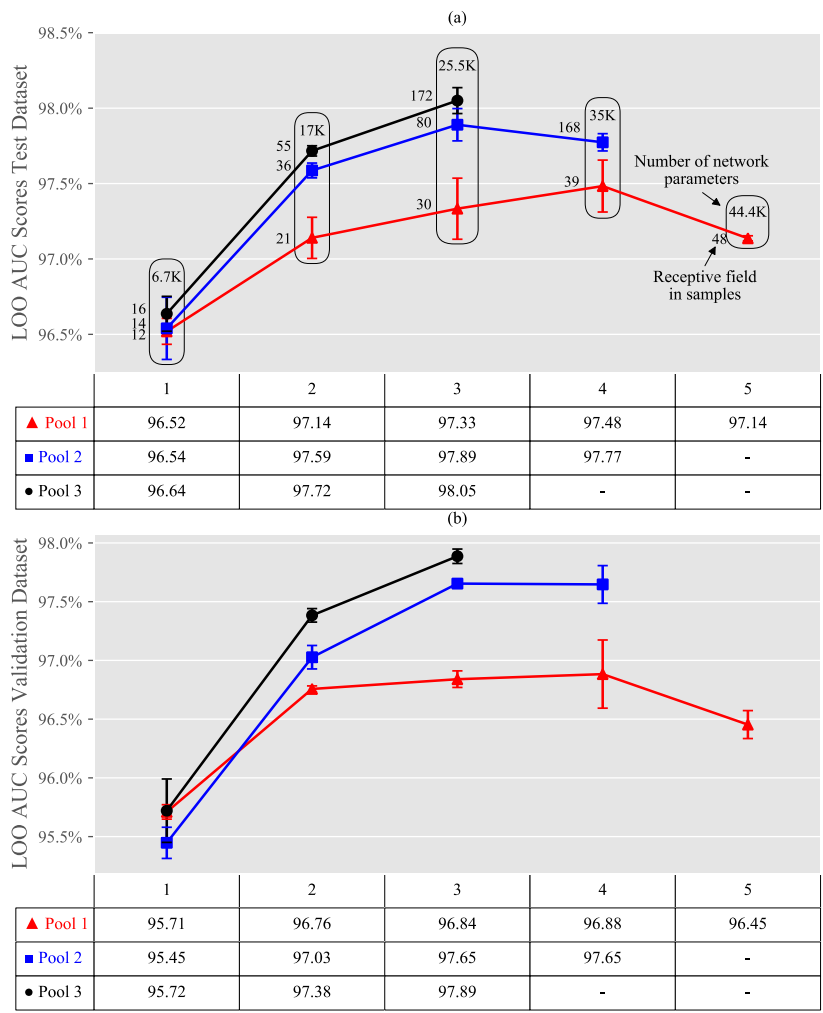

*Figure 7 The results of experiments comparing various combinations of 1D FCN architectural parameters. The horizontal axis represents the number of feature extraction blocks. The vertical axis represents AUC performance. The three different marker styles each represent a different value of pool stride. In (a) every point has a label that shows the maximum receptive field value for that network architecture configuration. Networks with the same number of trainable parameters are grouped together. (a) shows the performance on the test dataset at the end of the LOO experiment. (b) shows the corresponding performance on the validation dataset. For speed-up purposes validation results (b) exclude time-consuming post-processing steps whereas test results (a) include post-processing. For this reason, higher values of performance are observed on test data (a) as compared to validation data (b).*

*Table 2 Mean and standard deviation of AUC scores across the 18 babies in the Cork dataset. The Cork SVM which was developed in [25] is compared with the two classifiers developed in this work.*

|  |  | Cork SVM | 1D FCN | 2D FCN |
|---|---|---|---|---|
| AUC (%) | mean | 96.6 | 98.1 | 98.5 |
|  | std | 2.5 | 1.4 | 1.1 |
| AUC90 (%) | mean | 82.9 | 86.9 | 88.2 |
|  | std | 8.4 | 7.5 | 8.1 |

The pool stride is varied from one to three. Each value of pool stride is represented in Figure 7 using a different marker. The vertical axis represents the performance measures as AUC scores. All experiments were run three times starting from a different random initialisation point. Each point is the mean of three full LOO experiments, the error bars are the standard deviation of the experiments. The combination of the number of feature extraction blocks and the pool stride results in different receptive field widths and different numbers of total parameters for each architecture. In Figure 7 (a), each data point is labelled with the value of the receptive field in the final convolutional layer for that architecture. The receptive field increases with added convolutional depth and with the increased pool stride. The number of network parameters is not affected by the pool stride, so points with the same number of feature extraction blocks are grouped by number of parameters. Figure 7 (a) is the plot of mean AUC scores on the 18 held-out test babies. Figure 7 (b) shows the corresponding AUC scores on the validation data for each of the experiments in Figure 7 (a). The test AUC in Figure 7 (a) is calculated after the full post-processing routine is applied to the probabilistic output.

Table 2 shows the mean AUC and AUC90 LOO scores for the SVM Cork algorithm, the 1D FCN and the 2D FCN algorithm that correspond to the best performing architecture in Figure 7 (on the validation data). The SVM and 1D FCN are trained using the same subset of the Cork dataset which has channel-specific annotations and the 2D FCN is trained using the entirety of the Cork dataset (available for training after the LOO split). Table 2 also shows the standard deviations of the scores across the 18 babies in each LOO experiment. The relative performance improvement (achieved improvement over achievable improvement) using deep learning with respect to the Cork SVM baseline is 56% (98.5-96.6)/(100-96.6).

*Table 3 Comparison of LOO performance and held out test set performance for the 2D FCN algorithm trained on the Cork dataset, which is the dataset used in this work, and the SVM algorithm trained on the Helsinki dataset. The AUC results of training on Cork or Helsinki data and then testing on each dataset are shown. The test type, either Leave One patient Out or a once-off unseen test, is reported in brackets alongside each score.*

|  | AUC (%) | |
|---|---|---|
|  | **Test Database** | |
| **Algorithm** | Cork Dataset | Helsinki Dataset* |
| Helsinki SVM | (Test) 89.4 | (LOO) 95.5** |
| 2D FCN | (LOO) 98.5 | (Test) 95.6 |

\* AUC is reported on concatenated data so that the all babies can be included in the calculation
\*\* As reported in [13]

The performance of the deep learning algorithms on the publicly available Helsinki dataset and a comparison with the publicly available Helsinki algorithm are reported in Table 3. The 2D FCN which was trained using weakly labelled data on the Cork dataset, was shown to have the best performance on both the Cork dataset (LOO) and the unseen Helsinki test dataset.

## 4. Discussion

### 4.1 Comparison with the SVM baseline on Cork dataset

A direct comparison between the Cork SVM based system and the 1D FCN based system can be performed, as both systems use the same strongly labelled training data. Both systems operate on a single EEG channel at a time, share the same pre-processing and post-processing routines, and are tested in the same LOO fashion on the same dataset. The results indicate that the 1D FCN outperforms the SVM based system in the LOO experiment; the average performance of the 1D FCN shows an improvement of over 1% absolute, when compared to experiments with the Cork SVM. These results have proven the merits of exploring deep learning for the neonatal seizure detection task and have driven this work, which is an exploration of how far deep learning can improve the performance.

The inter-observer agreement is an important aspect to consider when working in the area of neonatal seizure detection [36]. No two neonatal electro-encephalographers will annotate long-term EEG recordings with 100% agreement. For an automated seizure detection algorithm to be clinically useful, its level of performance should match that of a human expert annotator i.e. it should have a comparable inter-observer agreement when compared with human experts.

The fact that a deep learning architecture has outperformed the SVM algorithm in Figure 6 shows that the convolutional layers can extract task appropriate features from the raw EEG signal without the need for hand-crafted features. While the achievements of deep learning algorithms in other signal processing fields certainly created the premise for this study, the ability for the deep learning algorithm to achieve a better AUC score than the highly engineered and tuned SVM algorithm required a rigorous process of architectural design. The magnitude of the increase achieved is however, very interesting, considering the years of engineering and medical expertise that went into the development and selection of the features in the SVM algorithm.

Neonatal EEG is a highly complex signal and it varies widely between babies. Many advancements have been made in the area of feature extraction from neonatal EEG signals. This had led to the development of feature sets which are optimised for the seizure detection task. Although these feature sets are task specific, it should be noted that whenever an intermediate representation of EEG is extracted, some of the signal information contained within the EEG waveform may be lost. The deep learning approach in this paper uses the temporal EEG signal; all of the information contained in the EEG is used to train the neural network. The fact that the FCNs have outperformed the state of the art algorithm on this large clinically representative dataset proves that the deep learning algorithm was able to extract more meaningful representations from raw EEG than can be done manually when restricted by prior knowledge and expert assumptions, such as signal stationarity (e.g. the application of the discrete Fourier transform). The engineer or the clinical expert is using their experience to decide which parts of the data will give the most information when choosing a feature set.

In contrast, the FCN has an entirely data-driven and objective-driven feature selection routine, as it is part of the backpropagation process. A data-driven latent representation is richer than the one that can be derived with human influence and intervention (inductive bias). An additional advantage is that both the feature extraction process and the input classification process are optimised in one end-to-end routine. The SVM algorithm is optimised in two stages, first the 55 features are designed and selected to be relevant for the seizure detection task, and next the classification algorithm is trained and optimised to distinguish between seizure and background EEG.

The FCN architecture offers numerous benefits when compared to a CNN that relies on fully connected layers. First, the number of network parameters are notably reduced, which along with weight sharing eliminates the need for regularisation routines such as dropout or L1-L2 regularisations. The absence of a dense layer allows the network to classify inputs of any length, since the last global average pooling layer can process a sequence of any length giving a constant sized output. While this is not explored in this study, this property will facilitate integration of the whole post-processing into a single optimisation routine. Finally, yet importantly, since only convolutional layers are used, the temporal relationship between the input and the output is preserved. In computer vision, FCN architectures have been used for object detection and localisation tasks with state-of-the-art results [37], [38]. In this work the challenges associated with deep learning such as large computational resources and the large amounts of data in the training stage, were overcome by training on a GPU and using a training dataset of over 800 hours of multi-channel EEG.

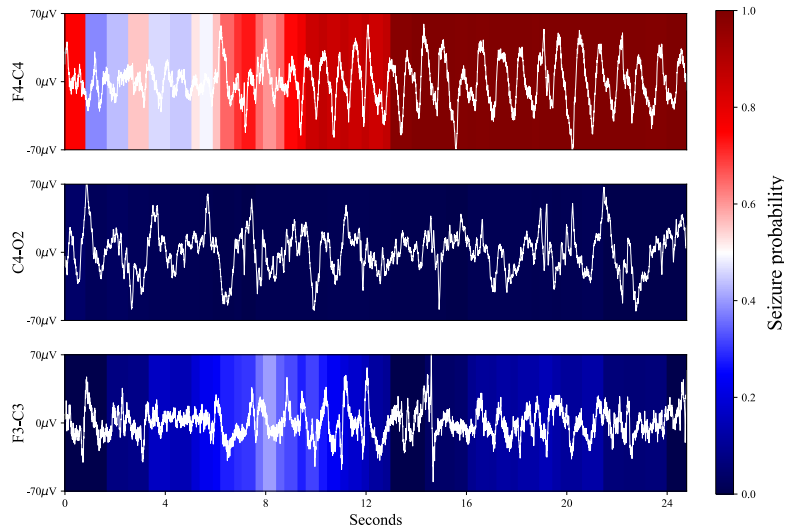

*Figure 8 A heatmap showing the probability of seizure for a 25 second sample of EEG over three channels. The EEG sample represented here is taken from the seizure event shown in Figure 1(a).*

Maintaining ordering in time throughout the classification network means that the input activity which maximally activates the feature maps in the FCN can be reviewed. Sections of EEG that contribute to seizure decisions can be investigated further by analysing the output of the final convolutional layer as a heatmap. Figure 8 shows a heatmap where the background corresponds to the level of seizure-ness that the convolutional network attributes to different parts of the input EEG signal. The colours shown in the background of this plot represent values in the output of the final convolutional layer. A softmax function was applied to corresponding values in the final two feature maps, giving values between zero and one, where higher values represent patterns which contributed highly to a seizure decision. Studying the EEG patterns that contribute to seizure decisions in this deep learning algorithm, could provide clinical insights into important characteristics of neonatal seizures.

## 4.2 Receptive field

The results reported for the deep learning algorithm in Figure 6 were the result of an architecture selection routine. This architecture selection routine leveraged the work of previous studies in this area to select a fully convolutional network over other varieties of CNN [15], [39]. This exploration of architectures also looked at varying the number of convolutional blocks, the stride of the pooling layers and the overall number of free parameters to tune.

Figure 7 (a) shows that increasing the number of network parameters does not lead to a guaranteed performance improvement. Similarly, the largest number of building blocks does not result in the highest score. At very low receptive field values, too much network complexity has resulted in a decrease in classifier performance; this was observed for experiments with a pool stride of one and the number of feature extraction blocks set to five. This specific experimental setup had the largest number of network parameters, but it produced a low performance compared to the other depths investigated.

Most of the improvement in AUC performance in this work is achieved by using wider receptive fields. The receptive field can be increased through adding more feature building blocks layers or by using a down-sampling operation (pooling). Figure 7 (a) clearly shows the importance of the receptive field width on the classifier's performance.

Originally, the FCN was developed for the task of semantic segmentation of images [40], and in this field, the effects of receptive field width have been studied in more depth. The success of the network developed in [41] is due to the inclusion of dilated convolutions. Dilated convolutions give an increase in receptive field without losing resolution or coverage, two very important considerations when trying to densely predict labels. Furthermore, the inclusion of one-sided dilated convolutions has been shown to outperform traditional recurrent neural networks in a series of sequence modelling tasks in [42]. Those results indicated that the influence of the receptive field in convolutional architectures is more valuable than the architectural choice of networks specifically designed for sequence modelling tasks. In this work, the importance of the receptive field for a fully convolutional temporal signal classification algorithm was explored; it is the first time that this architecture has been proposed for use with neonatal EEG signals.

It can be seen from Figure 7 (b) that the trends seen in the validation results reflect what is seen in the test results. This implies that the choice of architecture, the depth, the pool stride, etc, can be reliably selected using the validation dataset. When designing algorithms through a LOO process it is fundamental that the architecture selection and tuning routines are not informed by the performance of the algorithm on the held-out test baby but rather driven by the performance on the validation data. To avoid leakage between the training and testing sets a validation set is used to select the best hyper-parameters and dictate when to do early stopping. The FCN architectures (1D FCN and 2D FCN) used in LOO experiments in this work (Figure 6 and in Table 2) were chosen based on the best results achieved on the validation data in Figure 7 (b).

It is also important to highlight that the results in this section are not an exhaustive search of all possible architectures for the FCN. In Figure 7, a larger receptive field could be obtained by creating a more flexible building block or combining building blocks with different pool strides. The results presented here aim to show the importance of understanding, when designing a CNN architecture for a new data domain, which concepts matter, i.e. the receptive field, and which matter less, i.e. the depth or the number of parameters.

A recent publication which looks at the effects of scaling architectural parameters in a convolutional neural network within a fixed resource budget has shown state-of-the-art results on some open-source image recognition datasets [43]. This work varies three parameters, depth, width and input resolution, using a compounding scaling routine. The parameter investigation done in this study applies similar principles regarding network parameter balancing to networks designed for physiological signal processing.

## 4.3 More data with weak labels vs. less data with strong labels

The value of weakly labelled EEG signals is explored in this work by comparing the performance of the 1D FCN and 2D FCN. These networks have the same architectural parameters, the only difference is that the 2D FCN has 2D inputs and it includes a max pooling layer across the channel-dimension in the classification block. The annotations used in training the 2D FCN are weak in the sense that the network only knows if a seizure is present in at least one of the channels but has no channel-specific labels. The need for channel-specific labels when training machine learning models for the neonatal seizure detection task has been a bottleneck for performance improvement for many years. The need for annotators to provide channel-specific labels is time-consuming and often only small subsets of datasets have channel specific seizure labels. In this experiment, the comparison can be made between a deep learning model which was given channel-specific labels and a model which had weakly labelled seizures. Similar problems are faced in semantic segmentation for images, where the image has a label, but the individual pixels are not annotated and in the task of audio event detection [44].

The superior performance of the 2D algorithm, which was trained with weak labels, is because it was trained with over 40 times more EEG samples. Within each individual 2D training example, the increase in information due to fact that 8 channels are used instead of 1 might not be beneficial because 2D FCN sees more noise as the signal to noise ratio decreases. However, with the usage of weak labels the number of 2D examples increases considerably. The 1D FCN saw over half a million training examples in experiments of 8-seconds of single-channel EEG windows (~128M EEG samples). In contrast 2D FCN saw over 2.5 million training examples of 8-seconds of 8-channel EEG windows (~5B EEG samples). The ability to train with the whole training dataset, instead of just using a subset, shows the power of deep learning to deal with spatial and translational invariances and utilise higher amounts of data [45]. While the amount of data is highly important when training a deep learning model, it is worth noting that if strong labels were available for the whole dataset then the 1D FCN would not be outperformed by the 2D FCN architecture.

The 2D FCN's ability to learn from weakly labelled data opens the door for further improvement over neonatal seizure detection algorithms which require strongly labelled seizure data. If a 2D FCN, like the one designed in this work, was trained on an even larger dataset than the Cork dataset presented in this work the performance and generalisation ability could continue to improve. This would require further data collection and ethical approval, but it wouldn't require the labour intensive work of creating channel localised seizure annotations.

## 4.4 Comparison with a publicly available algorithm

The validation of an algorithm's performance on an unseen dataset is a good indicator of its generalisation ability. Table 3 illustrates the performances of the FCN developed in this work and the SVM Helsinki developed in [13], allowing the generalisation abilities of both algorithms to be compared in both matched and mismatched conditions. The matched condition is when the source of the dataset used for training and testing of the algorithm is the same. This implies that the properties of the dataset do not change from train to test – the variables such as the seizure burden, inter-observer variability in annotations, overall duration of recordings, montage setup, electrode application procedure, recording equipment characteristics with built-in proprietary filters, etc are kept constant. The mismatched condition occurs when these properties change between train and test. The 2D FCN has the best performance in both conditions. In addition, the drop in performance between matched and mismatched algorithm is smallest for 2D FCN – it achieves a much higher AUC performance on the Helsinki dataset than the Helsinki SVM algorithm achieves on the Cork dataset. All together it shows that the data-driven optimisation of feature extraction and classification is not only robust to these variabilities but also results in a more accurate classifier which has the ability to generalise across different data sources, different seizure and background characteristics.

The Helsinki dataset contains a large cohort, with EEG from 79 different babies. However, each baby is only represented by short duration EEG segments; on average 85 minutes. These short windows of EEG are not representative of the EEG recordings typically found in the NICU. Another important note is that the median postnatal age of the babies at the time of recording was three days. In babies who have experienced a hypoxic ischaemic event, most seizures begin in the first 24 hours after birth, and by day three the seizure burden has dropped by over 90% [46]. In the Cork dataset, used to train the FCNs in this work, the average postnatal age at recording start is less than 17 hours. The FCN algorithm was trained on long recordings of EEG with relatively low seizure activity; these conditions are typically present in the NICU. The post-processing which is included in the FCN algorithms was optimised with the typical NICU conditions in mind. The post-processing, which is described in more detail in [25], includes a moving average filter of 60 seconds and a long-term background adaptation.

The SVM Helsinki was trained on shorter EEG segments that had a relatively high seizure burden. The post-processing in this algorithm, which applies a decision window of 48-seconds, was tuned to deal with shorter recordings that have a higher proportion of seizure epochs. Both algorithms achieve an acceptable AUC value in their once-off unseen dataset tests, considering they were trained to deal with very different recording environments. The clinical differences between the dataset used in this work for training and for testing explain the variation between the algorithmic performances when tested on an unseen dataset of different clinical characteristics. The publication and release of publicly available dataset containing annotated neonatal seizures is an important step forward in this research field. However, a dataset with longer recordings would be more reflective of the situation that a seizure detection algorithm would face in the NICU.

It should be noted that there is a large variation between the processing time needed to implement each algorithm. The Helsinki SVM algorithm relies on the extraction of computationally intensive time-frequency features. In experiments, on the same PC and using the CPU for all computation, the Helsinki SVM takes over 4.5 hours to process one hour of EEG and the 1D FCN and the 2D FCN both take less than one minute to do the same task. When considering the clinical implementation of a neonatal seizure detection algorithm the processing needs to be realizable in real-time. In our previous work, a fully convolutional architecture was implemented on an Android tablet using Java [24] which indicates that such a system can provide real-time decision support in NICU even on a mobile device.

# 5. Conclusions

This work has presented a novel way of developing a neonatal seizure detection algorithm through end-to-end optimisation of the feature extraction and classification. The chosen deep learning algorithm is a fully convolutional neural network which operates on raw EEG signals; the use of this architecture is novel in the area of neonatal physiological signal processing. First, it has been shown that given the same initial conditions the 1D FCN algorithm, which learns a representation from large amounts of raw annotated data, outperforms an SVM-based system, which uses a set of hand-crafted features. Second, it has been shown, by varying the architectural parameters of the FCN that control the complexity and size of the internal learning representation, that the width of the receptive field in the final convolutional filters has the biggest influence on network performance when compared to the network depth or overall number of tuneable parameters. Finally, it has been shown that the designed novel 2D FCN architecture has achieved superior performance by exploiting much larger amounts of data with weak labels. By designing an algorithm which can learn from weakly labelled data this work hopes to accelerate the development of algorithms in the area of neonatal EEG by circumventing the need for laborious precise per-channel labelling of datasets.

The performance of the 2D FCN algorithm on the publicly available Helsinki dataset has indicated its superior performance to existing published alternatives. The performance reported in this work shows that an algorithm based only on deep learning is capable of outperforming algorithms that rely on complex feature sets without the computational load typically associated with intensive feature calculations. Future work in this area will focus on analysis of patterns learnt from large cohort of babies with long-term EEG recordings.

## Acknowledgements

The work was supported by Science Foundation Ireland Research Centre Award, INFANT (12/RC/2272). This publication has emanated from research supported in part by a research grant from Science Foundation Ireland (SFI) under the Grant Number 15/RI/3239. We gratefully acknowledge the support of NVIDIA Corporation with the donation of the TitanX GPU used for this research.